\pdfoutput=1

\documentclass[11pt]{article}

\usepackage{acl}

\usepackage{times}
\usepackage{latexsym}

\usepackage[T1]{fontenc}

\usepackage[utf8]{inputenc}
\usepackage{graphicx}
\usepackage{amsmath}
\usepackage{booktabs}
\usepackage{arydshln}
\usepackage{multirow}
\usepackage{amsfonts}
\usepackage{algorithm}
\usepackage{algpseudocode}
\usepackage{microtype}

\title{Modeling Hierarchical Reasoning Chains by Linking Discourse Units and Key Phrases for Reading Comprehension}

\author{Jialin Chen\textsuperscript{1}, Zhuosheng Zhang\textsuperscript{2,3}, Hai Zhao\textsuperscript{2,3,\thanks{\; Corresponding author. This work was supported in part by the Key Projects of National Natural Science Foundation of China under Grants U1836222 and 61733011.}} \\
 $^1$School of Mathematical Sciences, Shanghai Jiao Tong University \\
  $^2$Department of Computer Science and Engineering,  Shanghai Jiao Tong University\\ 
  $^3$Key Laboratory of Shanghai Education Commission for Intelligent Interaction \\
and Cognitive Engineering, Shanghai Jiao Tong University \\
\texttt{sjtuchenjl@sjtu.edu.cn, zhangzs@sjtu.edu.cn, zhaohai@cs.sjtu.edu.cn} \\}

\begin{document}
\maketitle
\begin{abstract}
Machine reading comprehension (MRC) poses new challenges over logical reasoning, which aims to understand the implicit logical relations entailed in the given contexts and perform inference over them. Due to the complexity of logic, logical relations exist at different granularity levels. However, most existing methods of logical reasoning individually focus on either entity-aware or discourse-based information but ignore the hierarchical relations that may even have mutual effects. In this paper, we propose a holistic graph network (HGN) which deals with context at both discourse level and word level, as the basis for logical reasoning, to provide a more fine-grained relation extraction. Specifically, node-level and type-level relations, which can be interpreted as bridges in the reasoning process, are modeled by a hierarchical interaction mechanism to improve the interpretation of MRC systems. Experimental results on logical reasoning QA datasets (ReClor and LogiQA) and natural language inference datasets (SNLI and ANLI) show the effectiveness and generalization of our method, and in-depth analysis verifies its capability to understand complex logical relations.
\end{abstract}
\section{Introduction}
Machine reading comprehension (MRC) is a challenging task that requires machines to answer a question according to given passages  \cite{hermann2015teaching,rajpurkar-etal-2016-squad,Rajpurkar2018Know,lai2017race}. A variety of datasets have been introduced to push the development of MRC to a more complex and more comprehensive pattern, such as conversational MRC \cite{reddy2019coqa,choi2018quac}, multi-hop MRC \cite{yang2018hotpotqa}, and commonsense reasoning \cite{davis2015commonsense,bhagavatula2019abductive,talmor2019commonsenseqa,huang2019cosmos}. In particular, some recent multi-choice MRC datasets pose even greater challenges to the logical reasoning ability of models \cite{yu2020reclor,ijcai2020-0501} which are not easy for humans to do well, either. Firstly, all the supporting details needed for reasoning are provided by the context, which means there is no additional commonsense or available domain knowledge. Secondly, it is a task of answer selection rather than answer retrieval, which means the best answer is chosen according to their logical fit with the given context and the question, rather than retrieved directly from the context according to the similarity between answers and context. Most importantly, the relations entailed in the contexts are much more complex than that of previous MRC datasets owing to the complexity of logic, which is hard to define and formulate. Without a targeted design for those challenges, existing pre-trained models, e.g., BERT, RoBERTa, fail to perform well in such kind of logical reading comprehension systems \cite{yu2020reclor,ijcai2020-0501}.

\begin{figure}
    \includegraphics[width=0.5\textwidth]{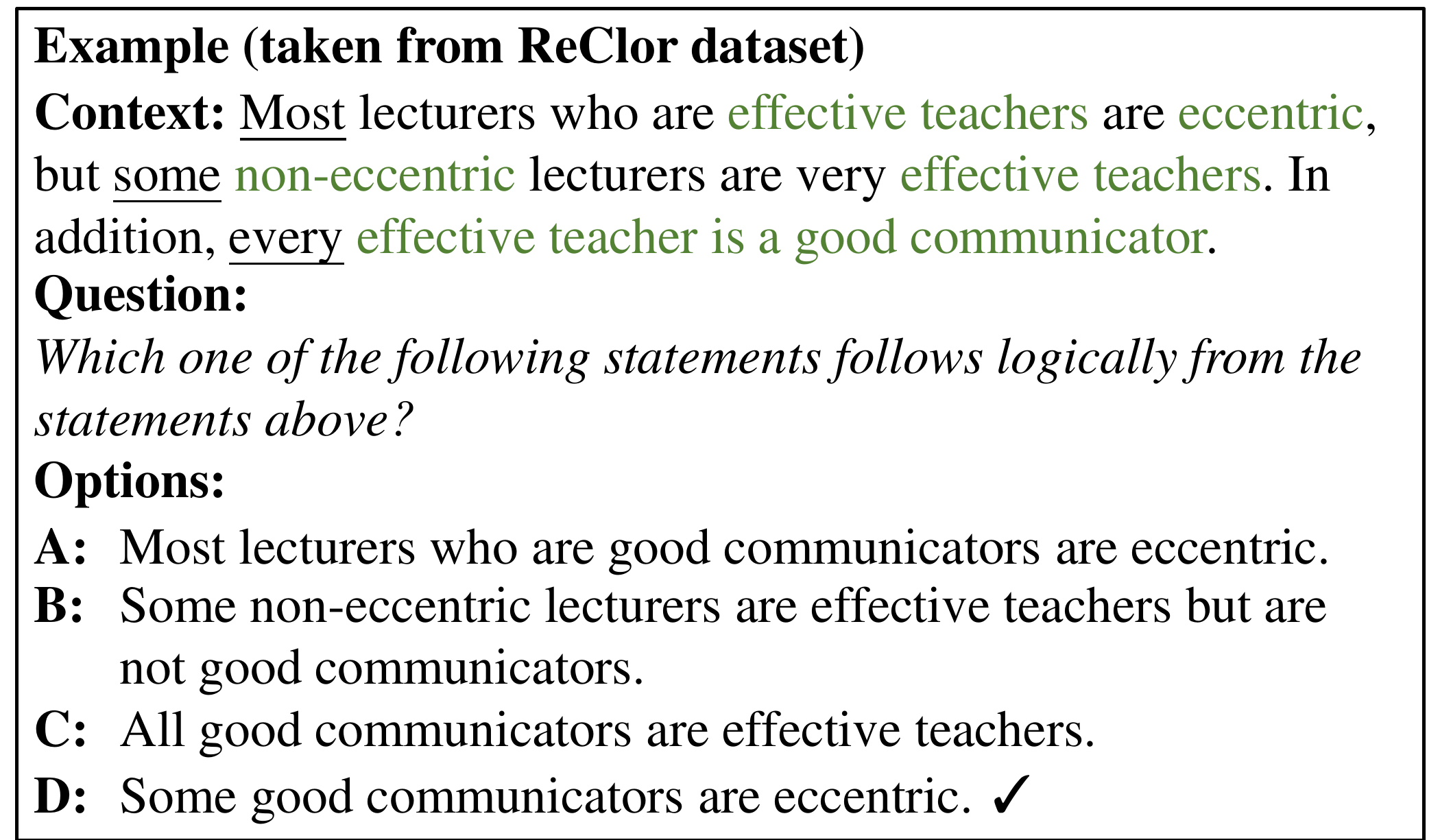}
    \caption{An example from Reclor dataset. The example mainly talks about "effective teachers, non-eccentric, eccentric, good communicator". }
    \label{examlple}
\end{figure}
\begin{figure*}[tp]
    \centering
    \includegraphics[width=0.95\textwidth]{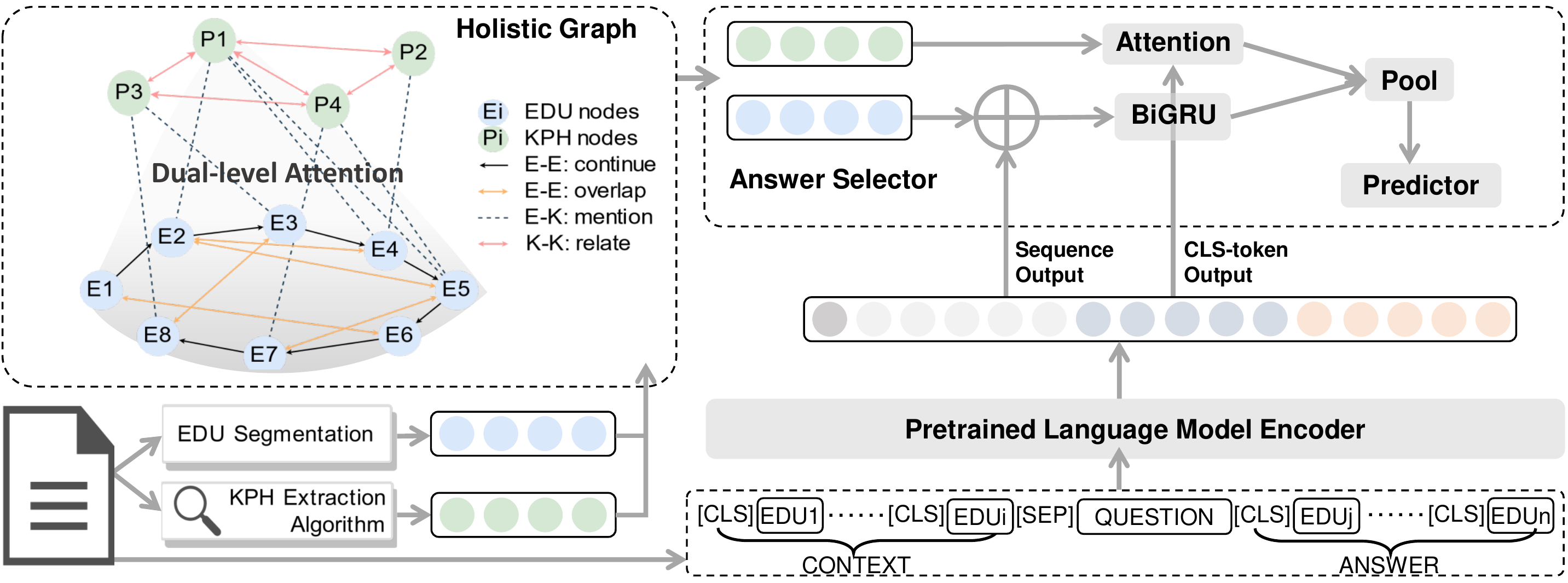}
    \caption{An overview of our proposed holistic graph-based reasoning model.}
    \label{model}
\end{figure*}
Logical reasoning MRC tasks are usually to find an appropriate answer, given a set of context and question. Figure \ref{examlple} shows an example from ReClor dataset \cite{yu2020reclor} which requires logical reasoning ability to make the correct predictions. As humans, to solve such problems, we usually go through the following steps. Firstly, we divide the context into several fragments and figure out the logical relations between each clause, such as transition, continuity, contrast, etc. 
Secondly, we extract the important elements in the context, namely, the objects and topics described by the context, and construct the logical graph with these significant elements. Finally, we need to compare the answer statement to the mentioned part in the context and assess its logical fit with the given context.

Most existing methods of logical reasoning MRC focus on either entity-aware or discourse-based information but ignore the hierarchical relations that may have mutual effects \cite{yu2020reclor,ijcai2020-0501,wang2021logic,zhang2021video,ouyang2021fact}. Motivated by the observation above, we model logical reasoning chains based on a newly proposed holistic graph network (HGN) that incorporates the information of element discourse units (EDU) \cite{gao-etal-2020-discern,ouyang2021dialogue} and key phrases (KPH) extracted from context and answer, with effective edge connection rules to learn both hierarchical features and interactions between different granularity levels. 

Our contributions are summarized as follows. (1) We design an extraction algorithm to extract EDU and KPH elements as the critical basic for logical reasoning. (2) We propose a novel holistic graph network (HGN) to deal with context at both discourse and word level with hierarchical interaction mechanism that yields logic-aware representation for reasoning. (3) Experimental results show our model's strong performance improvements over baselines, across multiple datasets on logical reasoning QA and NLI tasks. The analysis demonstrates that our model has a good generalization and transferability, and achieves higher accuracy with less training data.


\begin{figure*}[htp]
    \centering
    \includegraphics[width=0.95\textwidth]{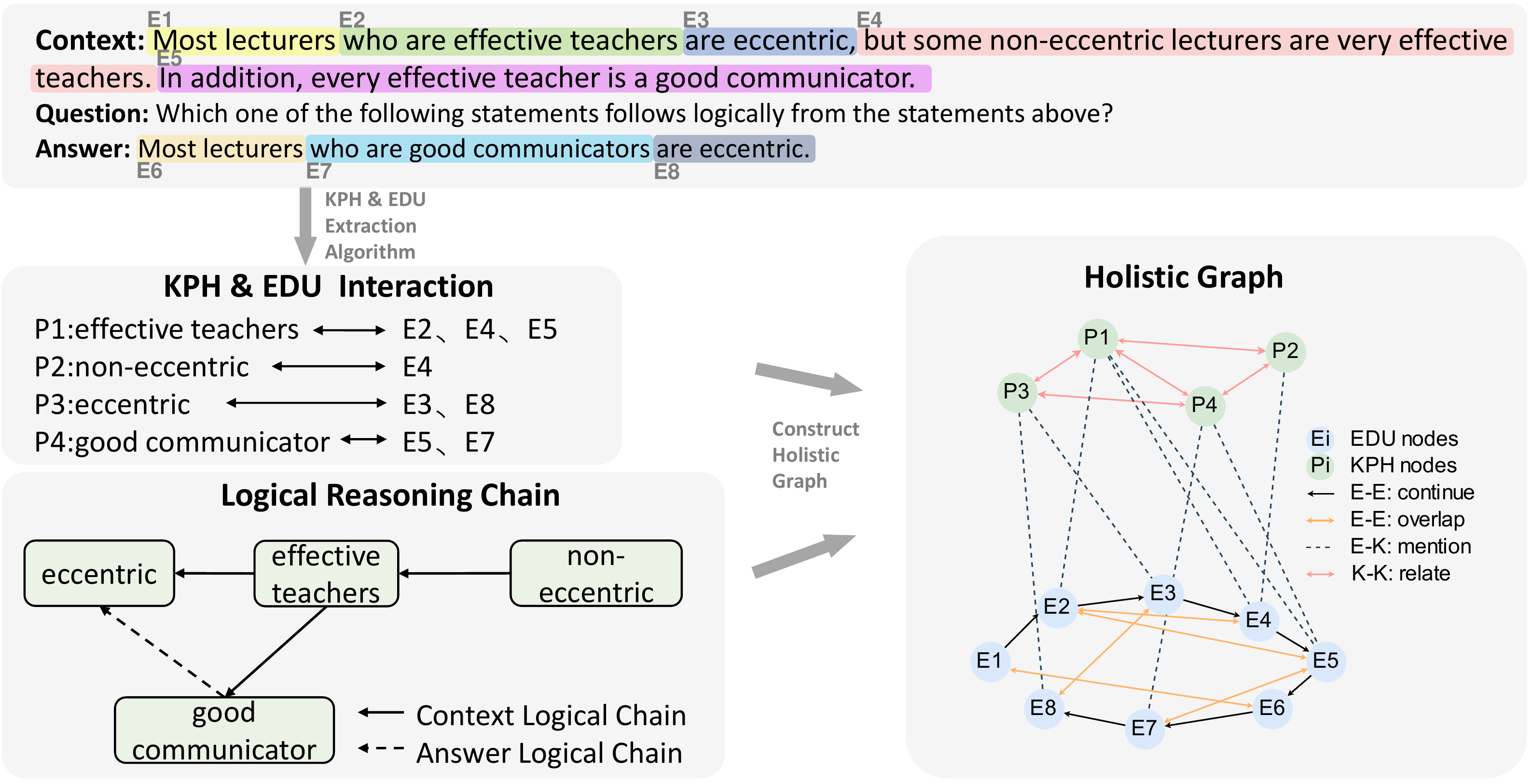}
    \caption{Process of constructing the holistic graph, using KPH-EDU Interaction information and pre-defined rules.}
    \label{process}
\end{figure*}
\section{Methodology}
Logical reasoning MRC tasks aim to find the best answer among several given options based on a piece of context that entails logical relations. Formally, given a natural language context \textsl{C}, a question \textsl{Q}, and four potential answers \textsl{A}=$\{A_1,A_2,A_3,A_4\}$. We concatenate them as $\{\textsl{C},\textsl{Q},A_i\}$ pairs. To incorporate the principle of human inference into our method, we propose a holistic graph network (HGN) as shown in Figure \ref{model}. Our model works as follows. First, we use EDU and KPH extraction algorithm to get necessary KPH nodes ($\{P_j\}$) and EDU nodes ($\{E_j\}$) from the given pairs. They contain information with different granularity levels and complement each other. Based on the extracted KPH-EDU interaction information and pre-defined rules, we construct the holistic graph. The process of constructing the holistic graph is shown in Figure \ref{process}. Then we measure the interaction between $\{E_j\}$ and $\{P_j\}$ to obtain logic-aware representations for reasoning.

\subsection{Logical Chain Construction}
\paragraph{Element Discourse Units (EDU)}
We use clause-like text spans delimited by logical relations to construct the rhetorical structure of texts. These clause-like discourses can be regarded as element units that reveal the overall logic and emotional tone of the text. For example, conjunctions like "\verb+because+'' indicate a causal relation which means the following discourse is likely to be the conclusion we need to pay attention to. Parenthesis and clauses like "\verb+who are effective teachers+" in Figure \ref{process} play a complementary role in context. Also, punctuation indicates a pause or an end of a sentence, containing semantic transition and turning point implicitly. We use an open segmentation tool, SEGBOT \cite{li2018segbot}, to identify the element discourse units (EDUs) from the concatenation of \verb+context+ and \verb+answer+, ignoring the question whose structure is simple. Conjunctions (e.g., "\verb+because+", "\verb+however+"), punctuation and the beginning of parenthesis and clauses (e.g., "\verb+which+", "\verb+that+") are usually the segment points. They are considered as explicit discourse-level logical relations.

To get the initial embedding of EDUs, we insert an external
\verb+[CLS]+ symbol at the start of each discourse, and add a \verb+[SEP]+ symbol at the end of every type of inputs. Then we use RoBERTa to encode the concatenated tokens. The encoded \verb+[CLS]+ token represents the following EDU. Therefore, we get the initial embedding of EDUs.

\paragraph{Key Phrase (KPH)}
\label{gram}
Key Phrases, including keywords here, play an important role in context. They are usually the object and principle of a context. We use the sliding window to generate $n$-gram word list, filtering according to the Stopword list, POS tagging, the length of the word, and whether it contains any number.\footnote{The stop list is derived by the open-source toolkit Gensim: \url{https://radimrehurek.com/gensim/}. The POS tagging is derived by the open-source toolkit NLTK: \url{https://www.nltk.org/}.} The filtering process is based on the following two main criteria:

(1) If the $n$-gram contains a stop word or a number, then delete it. 

(2) If the length of word is less than the threshold value $m$, delete it, and if the $n$-gram length is 1, then only the noun, verb, and adjective are retained. 

Then, we calculate the TF-IDF features of each $n$-gram, and select the top-$k$ $n$-gram as key phrases. $k$ is a hyper-parameter to control the number of KPHs. We restore the selected tokens and retrieve the original expressions containing the key phrase from the original text. For example, as in Figure \ref{process}, "\verb+eccentric+" is one of the KPHs, while we retrieve the original expression "\verb+eccentric+" and "\verb+non-eccentric+" from the original text. \footnote{The complete algorithm is given in Appendix \ref{kph_algorithms}.} 

Given the token embedding sequence $K_i = \{t_1,\dots,t_n\}$ of a KPH with length $n$, its initial embedding is obtained by 
\begin{equation}
    P_i = \frac{1}{|K_i|}\sum_{t_l\in K_i}t_l.
\end{equation}

\paragraph{Holistic Graph Construction}
Formally, every input sample is a triplet that consists of a context, a question and a candidate answer. EDU and KPH nodes are extracted in the above way. As shown in Figure \ref{process}, 
we construct a holistic graph with two types of nodes: EDU Nodes (in blue) and KPH Nodes (in green). For edge connections, there are four distinct types of edges between pairs of nodes.

$\bullet$ EDU-EDU continue: the two nodes are contextually associated in the context and answer. This type of edge is directional.

$\bullet$ EDU-EDU overlap: the two nodes contain the same KPH. This type of edge is bidirectional.

$\bullet$ EDU-KPH mention: the EDU mentions the KPH. This type of edge is bidirectional.

$\bullet$ KPH-KPH relate: the two nodes are semantically related. We define two types of semantic relations. One is that the two KPHs are retrieved by the same $n$-gram as described above. The other one is that the Cosine similarity between the two KPH nodes is greater than a threshold. This type of edge is bidirectional and can capture the information of word pairs like synonyms and antonyms.

The construction of the graph is based on intuitive rules, which will not introduce extra parameters or increase model complexity. A further parameter comparison is given in Table \ref{complexity}.
\subsection{Hierarchical Interaction Mechanism}
Considering a specific node in the holistic graph, neighboring nodes in the same type may carry more salient information, thus affecting each other in a direct way. In the process, the neighboring nodes in the different types may also interact with each other. To capture both the node-level and type-level attention, we apply a Hierarchical Interaction Mechanism to the update of the graph network's representations.
\paragraph{Graph Preliminary}
Formally, consider a graph $G=\{V,E\}$, where $V$ and $E$ represent the sets of nodes and edges respectively. $A$ is the adjacency matrix of the graph. $A_{ij}>0$ means there is an edge from the $i$-th node to the $j$-th node. We introduce ${A}'=A + I$ to take self-attention into account. In order to avoid changing the original distribution of the feature when multiplying with the adjacency matrix, we normalize ${A}'$, set $\tilde{A} = D^{-\frac{1}{2}}{A}'D^{-\frac{1}{2}}$ where $D$ is the degree matrix of the graph. $D = diag\{d_1,d_2\dots,d_n\}$, $d_i$ is the number of edges attached to the $i$-th node. 

Now, we calculate the attention score from node ${v}'$ to node $v$ in the following steps.
\paragraph{Type Attention Vector}
We use $T(\tau)$ to represent all nodes that belong to type $\tau$, and $N(v)$ to represent all neighboring nodes that are adjacent to $v$. $T$ is the set of types. Assume that node $v$ belongs to $T(\tau)$, $h_\mu$ is the feature of node $\mu$, $h_\tau$ is the feature of type $\tau$ which is computed by 
\begin{equation}
h_\tau = \sum_{\mu\in T(\tau)}\tilde{A}_{v \mu}W h_{\mu}. 
\end{equation}
Using the feature of type and node $v$, we compute the attention score of type $\tau$ as:
\begin{equation}
    e_{\tau} = \sigma(\mu_\tau^{T}\cdot[Wh_v\parallel W_{\tau}h_{\tau}]).
\end{equation}
Then, type-level attention weights $\alpha_\tau$ is obtained by normalizing the attention scores across all the types $T$ with the softmax function. $\sigma$ is an activate function such as leaky-ReLU.
\begin{equation}
\alpha_\tau = \frac{\textup{exp}(\sigma(\mu_\tau^{T}\cdot[Wh_v\parallel W_{\tau}h_{\tau}]))}{\sum_{{\tau}' \in T} \textup{exp}(\sigma(\mu_{{\tau}'}^{T}\cdot[Wh_v\parallel W_{{\tau}}h_{{\tau}'}]))}.    
\end{equation}
\paragraph{Node Attention Vector}
$\alpha_\tau$ shows the importance of nodes in type $\tau$ to node $v$. While computing the attention score of node ${v}'$ that is adjacent to node $v$, we multiply that by the type attention weights $\alpha_{\tau}$ (assume ${v}'$ belongs to type $\tau$). Similarly, node attention weights are obtained by the softmax function across all neighboring nodes.
\begin{equation}
     e_{v{v}'} = \sigma(\nu^T\cdot \alpha_\tau[Wh_v\parallel Wh_{{v}'}]),
\end{equation}
\begin{equation}
    \alpha_{v{v}'}=\frac{\textup{exp}(e_{v{v}'})}{\sum_{i\in N(v)}\textup{exp}(e_{vi})},
\end{equation}
where $\parallel$ is the concatenation operator and $\alpha_{v{v}'}$ is the attention weight from node ${v}'$ to $v$.
\paragraph{Update of Node Representation}
Let $h_{v}^{(l)}$ be the representation of the node $v$ at the $l$-th layer. Then the layer-wise propagation rule is as follows:
\begin{equation}
    h_{v}^{(l+1)} = \sigma(\sum_{{v}'\in N(v)}\alpha_{v{v}'}Wh_{{v}'}^{(l)}).
\end{equation}

\subsection{Answer Selector}
To predict the best answer that fits the logic entailed in the context, we extract the node representations of the last layer of the graph network and feed them into the downstream predictor. For EDU nodes, since the node order implies the occurrence order in the context, we align them with the output of sequence embedding and add to it as a residual part. Therefore, we feed them into a bidirectional gating recurrent unit (BiGRU).
\begin{equation}
    \tilde{H}_{E} = \textup{BiGRU}(H_{E}+H_{sent})\in \mathbb{R}^{l\times d},
\end{equation}
where $H_{E} = [h_{{v}'_1},h_{{v}'_2},\dots,h_{{v}'_l}]\in \mathbb{R}^{l\times d}$, ${v}'_i$ belongs to type EDU. $l$ and $d$ are the sequence length and the feature dimension respectively. $H_{sent}$ is the output of sequence embedding.

For KPH nodes, we first expand the embedding of the first \verb+[CLS]+ token to size $1\times d$, denoted as $H_c$. Then, we feed the embedding of \verb+[CLS]+ token and features of KPH nodes $H_{K} = [h_{v_1},h_{v_2},\dots,h_{v_n}]\in \mathbb{R}^{n\times d}$ ($v_i$ is of KPH type) into an attention layer.
\begin{equation}
\begin{split}
    \alpha_i&=w_{\alpha}^T[H_{c}\parallel h_{v_i}]+b_\alpha \in \mathbb{R}^1,\\
    \tilde{\alpha_i}&=\textup{softmax}(\alpha_i) \in [0,1],\\
    \tilde{H}_c&=W_c\sum_i\tilde{\alpha_i}h_{v_i}+b_c \in \mathbb{R}^{1\times d},
\end{split}
\label{dec}
\end{equation} 
where $\tilde{\alpha_i}$ is the attention weight of node feature $h_{v_i}$. $w_{\alpha}$, $b_\alpha$, $W_c$, and $b_c$ are parameters.

The output of BiGRU and the output of attention layer are concatenated and go through a pooling layer, followed by an MLP layer as the predictor. We take a weighted sum of the concatenation as the pooling operation. The predictor is a two-layer MLP with a tanh activation. Specially, coarse-grained and fine-grained features are further fused here to extract more information.
\begin{equation}
        \tilde{H} = W_p[\tilde{H}_{E}\parallel\tilde{H}_c],\quad
        p =\textup{MLP}(\tilde{H})\in \mathbb{R},
\end{equation}
where $W_p$ is a learnable  parameter, $\parallel$ is the concatenation operator. For each sample, we get $P=[p_1,p_2,p_3,p_4]$, $p_i$ is the probability of $i$-th answer predicted by model.  

The training objective is the cross entropy loss: 
\begin{equation}
\mathcal{L} = -\frac{1}{N}\sum_{i}^{N}\log \textup{softmax}(p_{y_i}),
\end{equation}
where $y_{i}$ is the ground-truth choice of sample $i$. $N$ is the number of samples.

\begin{table*}[t]
\centering
\setlength{\tabcolsep}{6.8pt}
\begin{tabular}{lcccccc}
\toprule
\multirow{2}{*}{Model}& \multicolumn{4}{c}{ReClor} & \multicolumn{2}{c}{LogiQA} \\
& Dev & Test & Test-E & Test-H & Dev & Test\\ 
\midrule
Human $^{\diamondsuit}$ & - & 63.0 & 57.1 & 67.2 & - &86.0\\
\midrule
RoBERTa$_\textsc{base}$ $^\diamondsuit$ & 55.0 & 48.5 & 71.1& 30.7& 33.3$^\star$& 32.7$^\star$ \\
\textbf{HGN}$_\textsc{RoBERTa(B)}$&56.3$^{(\uparrow1.3)}$ &51.4$^{(\uparrow2.9)}$&75.2$^{(\uparrow4.1)}$& 32.7$^{(\uparrow2.0)}$&39.5$^{(\uparrow6.2)}$&35.0$^{(\uparrow2.3)}$\\
\hdashline
RoBERTa$_\textsc{large}$ $^\diamondsuit$ &62.6&55.6&75.5&40.0&35.0&35.3\\
DAGN $^\diamondsuit$ &65.2&58.2&76.1&44.1&35.5&38.7 \\
DAGN (Aug) $^\diamondsuit$  &65.8&58.3&75.9&44.5&36.9&39.3\\
LReasoner$_\textsc{RoBERTa}^\spadesuit$&66.2&62.4&81.4&47.5&38.1&40.6\\
$\quad$- data augmentation $^\spadesuit$& 65.2 &58.3& 78.6& 42.3& - &-\\
\textbf{HGN}$_\textsc{RoBERTa(L)}$&66.4$^{(\uparrow3.8)}$&58.7$^{(\uparrow3.1)}$&77.7$^{(\uparrow2.2)}$& 43.8$^{(\uparrow3.8)}$ &40.1$^{(\uparrow5.1)}$&39.9$^{(\uparrow4.6)}$\\
\hdashline
DeBERTa$^\spadesuit$ &74.4&68.9&83.4&57.5&44.4&41.5\\
LReasoner$_\textsc{DeBERTa}^\spadesuit$&74.6&71.8&83.4&\textbf{62.7}&\textbf{45.8}&43.3\\
\textbf{HGN}$_\textsc{DeBERTa}$&\textbf{76.0}$^{(\uparrow1.6)}$&\textbf{72.3}$^{(\uparrow3.4)}$&\textbf{84.5}$^{(\uparrow1.1)}$&\textbf{62.7}$^{(\uparrow5.2)}$&44.9$^{(\uparrow0.5)}$&\textbf{44.2}$^{(\uparrow2.7)}$\\
\bottomrule
\end{tabular}
\caption{Experimental results (Accuracy: $\%$) of our model compared with baseline models on ReClor and LogiQA datasets. Test-E and Test-H denote Test-Easy and Test-Hard subclass of the ReClor dataset respectively. The results in \textbf{bold} are the
best performance of all models. $\diamondsuit$ indicates that the results are given by \citet{zhang2021video}, $\spadesuit$ indicates the results are given by \citet{wang2021logic}, $\star$ means that the results 
come from our own implementation. $\textsc{roberta(L)}$ and $\textsc{roberta(B)}$ denotes RoBERTa-large and RoBERTa-base, respectively.
\label{table1}}
\end{table*}

\begin{table*}[t]
\centering
\setlength{\tabcolsep}{3.2pt}{
\begin{tabular}{lcccccccc}
\toprule
\multirow{2}{*}{\% data used}&\multicolumn{2}{c}{0.1\%}&\multicolumn{2}{c}{1\%}&\multicolumn{2}{c}{10\%}&\multicolumn{2}{c}{100\%}\\
&Dev&Test&Dev&Test&Dev&Test&Dev&Test\\
\toprule
BERT$_\textsc{base}$&73.2&70.4&77.9&76.8&84.2&83.9&90.8&90.7\\
RoBERTa$_\textsc{large}$&84.8&82.0&\textbf{87.6}&87.0&89.5&88.8&92.2&91.0\\
\textbf{HGN}$_\textsc{Bert(B)}$&75.8$^{(\uparrow 2.6)}$&75.4$^{(\uparrow 5.0)}$&81.1$^{(\uparrow 3.2)}$&80.3$^{(\uparrow3.5)}$&85.4$^{(\uparrow1.2)}$&83.9$^{(\uparrow0.0)}$&91.3$^{(\uparrow0.5)}$&91.0$^{(\uparrow0.3)}$\\
\textbf{HGN}$_\textsc{RoBERTa(L)}$&\textbf{85.4}$^{(\uparrow0.6)}$&\textbf{83.5}$^{(\uparrow1.5)}$&\textbf{87.6}$^{(\uparrow0.0)}$&\textbf{87.3}$^{(\uparrow0.3)}$&\textbf{90.2}$^{(\uparrow0.7)}$&\textbf{89.4}$^{(\uparrow0.6)}$&\textbf{92.3}$^{(\uparrow0.1)}$&\textbf{91.5}$^{(\uparrow0.5)}$\\
\bottomrule
\end{tabular}}
\caption{Experimental results (Accuracy: $\%$) on the SNLI dataset. We randomly generate the training dataset with limited size, without changing the size of Dev. and Test set. $\textsc{Bert(B)}$ and $\textsc{RoBERTa(L)}$ denote BERT-base and RoBERTa-large respectively.\label{snli}}
\end{table*}

\begin{table*}[t]
\centering
\setlength{\tabcolsep}{0.5mm}
\begin{tabular}{lcccccccc}
\toprule
\multirow{2}{*}{Model}& \multicolumn{4}{c}{Dev}&\multicolumn{4}{c}{Test}\\
&A1&A2&A3&ANLI&A1&A2&A3&ANLI\\
\toprule
RoBERTa$_\textsc{large}$&74.1&50.8&43.9&55.5&73.8&48.9&44.4&53.7\\
ALUM$^\spadesuit$&73.3&53.4&48.2&57.7&72.3&52.1&48.4&57.0\\
InfoBERT$^\diamondsuit$&76.4&51.7&48.6&58.3&75.5&51.4&49.8&58.3\\
\textbf{HGN}$_\textsc{RoBERTa(L)}$&\textbf{76.7}$^{(\uparrow2.6)}$&\textbf{69.3}$^{(\uparrow18.5)}$&\textbf{74.5}$^{(\uparrow30.6)}$&\textbf{71.3}$^{(\uparrow15.8)}$&\textbf{79.5}$^{(\uparrow5.7)}$&\textbf{63.4}$^{(\uparrow14.5)}$&\textbf{76.3}$^{(\uparrow31.9)}$&\textbf{68.9}$^{(\uparrow15.2)}$\\
\bottomrule
\end{tabular}
\caption{\label{anli}Experimental results (Accuracy: $\%$) on the ANLI dataset. Both ALUM and InfoBERT take RoBERTa-large as the backbone model. $\spadesuit$ means the results from \citet{liu2020adversarial}. $\diamondsuit$ means the results from \citet{wang2020infobert}.}
\end{table*}

\section{Experiment}

\subsection{Dataset}
Our evaluation is based on logical reasoning MRC benchmarks (ReClor \cite{yu2020reclor} and LogiQA \cite{ijcai2020-0501}) and natural language inference benchmarks (SNLI \cite{bowman2015large} and ANLI \cite{nie-etal-2020-adversarial}). ReClor contains 6,138 multiple-choice questions modified from standardized tests. LogiQA has more instances (8678 in total) and is derived from expert-written questions for testing human logical reasoning ability \cite{ijcai2020-0501}. To assess the generalization of models on NLI tasks, we test our model on the Stanford Natural Language Inference (SNLI) dataset, which contains 570k human annotated sentence pairs. The Adversarial Natural Language Inference (ANLI) is a new large-scale NLI benchmark dataset, where the instances are chosen to be difficult for the state-of-the-art models such as BERT and RoBERTa. It can be used to evaluate the generalization and robustness of the model.\footnote{The statistics information of these datasets are given in Appendix \ref{dataset}.}

Implementation details and parameter selection are reported in Appendix \ref{params} for reproduction.\footnote{Our source codes is available at \url{https://github.com/Cather-Chen/Logical-Reasoning-Graph}.}

%
\subsection{Main Result}
\subsubsection{Results on Logical QA}
Table \ref{table1} presents the detailed results on the development set and the test set of both ReClor and LogiQA datasets. We observe consistent improvements over the baselines. \textbf{HGN}$_\textsc{RoBERTa(B)}$ reaches $51.4\%$ of test accuracy on ReClor, and $35.0\%$ of test accuracy on LogiQA, outperforming other existing models. \textbf{HGN}$_\textsc{RoBERTa(L)}$ reaches $58.7\%$ of test accuracy on ReClor, therein $77.7\%$ on Easy subset and $43.8\%$ on Hard subset, and $39.9\%$ on LogiQA. \textbf{HGN}$_\textsc{DeBERTa}$ achieves $72.3\%$ on the test set of ReClor and $44.2\%$ on LogiQA. If using the same pre-trained language models as the backbones, our proposed model achieves the state-of-the-art results on both ReClor and LogiQA, without extra human annotations. Our model shows great improvement over this task by better utilizing the interaction information , which is ignored by most existing methods.

\subsubsection{Results on general NLI tasks}
To verify the generality of our model, we conduct experiments on two widely used entailment datasets for NLI: SNLI and ANLI, in which existing models rarely emphasized the modeling of logical relations. Table \ref{snli} compares the performances of \textbf{HGN} and baseline models on the SNLI dataset with the same proportion of training data for finetuning. We observe that when given a limited number of training data, our \textbf{HGN} has faster adaptation than baseline models as evidenced by higher performances in low-resource regimes (e.g., $0.1\%$, $1\%$, and $10\%$ of the training data used). \textbf{HGN} also outperforms BERT$_\textsc{base}$ by $0.3\%$ and RoBERTa$_\textsc{large}$ by $0.5\%$ on the full SNLI.
\begin{figure}[h]
    \includegraphics[width=0.45\textwidth]{{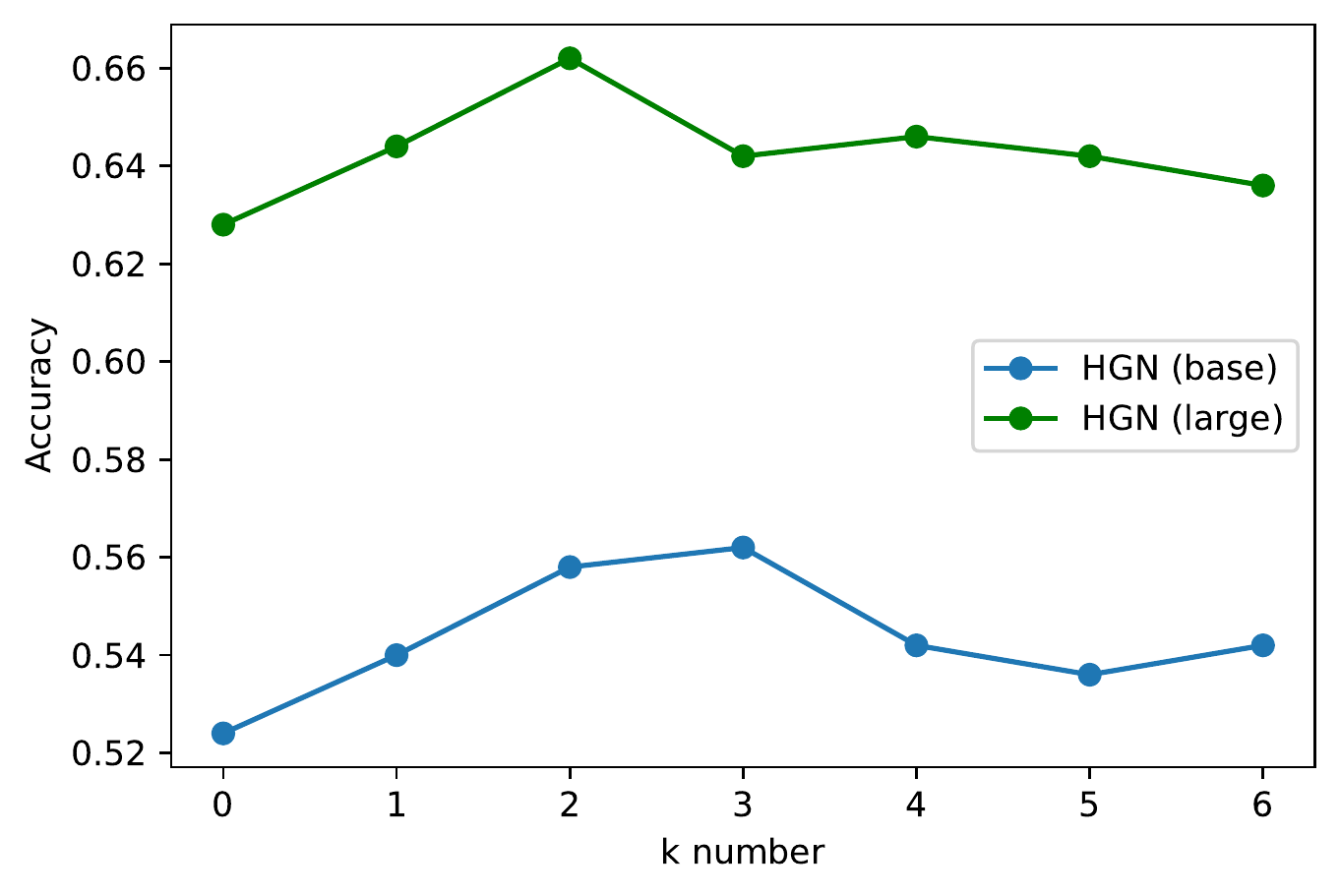}}
    \caption{Dev. accuracy on the ReClor dataset as the number of KPH nodes changes.\label{k}}
\end{figure}
We assess the model’s robustness against adversarial attacks, using a standard
adversarial NLP benchmark: ANLI, as shown in Table \ref{anli}. A1, A2 and A3 are three rounds with increasing difficulty and data size. ANLI refers to the combination of A1, A2 and A3. \textbf{HGN}$_\textsc{RoBERTa(L)}$ gains a $15.2\%$ points in test accuracy of ANLI over RoBERTa$_\textsc{large}$, creating state-of-the-art results on all rounds. Results show that our model has a comprehensive improvement over baseline models, in aspects of faster adaption, higher accuracy and better robustness.

\subsection{More Results}
\paragraph{Interpretation of $k$}
In this part, we investigate the sensitivity of parameter $k$, which is the number of KPH node. Figure \ref{k} shows the accuracies on the development set of our proposed model with different numbers of KPH nodes, which are extracted according to TF-IDF weights. We observe that $k=2$ or $k=3$ is an appropriate value for our model. This is consistent with our intuition that a paragraph will have 2 to 3 key phrases as its topic. When $k$ is too small or large, the accuracy of the model does not perform well.

\paragraph{Model Complexity}
With well-defined construction rules and an appropriate architecture, our model enjoys the advantage of high performance with fewer parameters. We display the statistics of model's parameters in Table \ref{complexity}. Compared with the baseline model (RoBERTa$_\textsc{large}$), the increase of our model's parameters is no more than $4.7\%$. Particularly, our model contains fewer parameters and achieves better performance than DAGN.
\subsection{Ablation}

\begin{table}[t]
\centering
\setlength{\belowcaptionskip}{-0.45cm}
\setlength{\tabcolsep}{4pt}{
\begin{tabular}{l|ccc}
\toprule
Model& RoBERTa&DAGN&\textbf{HGN}\\
\hline
Params& 356.4M &396.2M &373.4M\\
\bottomrule
\end{tabular}}
\caption{Statistics of models' parameters\label{complexity}}
\end{table}

We conduct a series of ablation studies on Graph Construction, Hierarchical Interaction Mechanism and Answer Selector. Results are shown in Table \ref{ablation}. All models use RoBERTa-base as the backbone.
\paragraph{Holistic Graph Construction}
The Holistic Graph in our model contains two types of nodes and four types of edges. We remove the nodes of EDU and KPH respectively and the results show that the removal hurts the performance badly. The accuracies drop to $55.8\%$ and $53.9\%$. Furthermore, we delete one type of edge respectively. The removal of edge type destroys the integrity of the network and may ignore some essential interaction information between EDUs and KPHs, thus causing the drop of the performance.
\paragraph{Hierarchical Interaction Mechanism}
Hierarchical Interaction Mechanism helps to capture the information contained in different node types. When we remove the type-level attention, the model is equivalent to a normal Graph Attention Network (GAT), ignoring the heterogeneous information. As a result, the performance drops to $54.8\%$. When we remove both types of attention, the performance drops to $55.7\%$.
\begin{table}[t]
\centering
\setlength{\tabcolsep}{1.6pt}
\begin{tabular}{ll}
    \toprule
    Model    & Accuracy ($\%$) \\
    \midrule
    \textsc{HGN}$_\textsc{base}$ & \textbf{56.3}\\
    \midrule
    \textit{Graph Construction}  \\
    \quad - EDU  & 55.8 (↓0.5)\\
    \quad - KPH & 53.9 (↓2.4)\\
    \quad - edge type: E-E continue & 53.0 (↓3.3)\\
    \quad - edge type: E-E overlap &  54.0 (↓2.3)\\
    \quad - edge type: E-K mention & 54.2 (↓2.1)\\
    \hdashline
    \textit{Hierarchical Interaction} \\
    \quad - type-level attention (i.e. GAT) & 54.8 (↓1.5)\\
    \quad - both (i.e. GCN) & 55.7 (↓0.6)\\
    \hdashline
    \textit{Answer Selector} \\
    \quad - BiGRU     &  53.2 (↓3.1) \\
    \quad - Attention layer     & 55.0 (↓1.3)\\
    \bottomrule
  \end{tabular}
  \caption{Ablation results on the dev set of ReClor.}\label{ablation}
\end{table} 
\paragraph{Answer Selector}
We make two changes to the answer selector module: (1) deleting the BiGRU, (2) deleting the attention layer. For (1), the output of EDU features concatenates with the output of the attention layer directly and then are fed into the downstream pooling layer. For (2), we ignore the attention between the KPH features and the whole sentence-level features. The resulting accuracies of (1) and (2) drop to $53.2\%$ and $55\%$, which verify that the further fusion of features with different granularity is necessary in our proposed model.

We further analysed the examples that are predicted correctly by our model but not by baselines, and found that the powerful pre-trained language models, such as RoBERTa, would bias for answers with higher similarity to the context or those containing more overlapping words. The model itself does not understand the logical relations, but only compares their common elements for prediction. Instead, our model can not only match synonymic expressions, but also make logical inferences by separating sentences into EDUs and extracting key phrases and establishing logical relations between them. An example is shown in Appendix~\ref{appendix:case}.
\section{Related Work}
\subsection{Machine Reading Comprehension}
MRC is an AI challenge that requires machines to answer questions based on a given passage, which has aroused great research interests in the last decade \cite{hermann2015teaching,rajpurkar-etal-2016-squad,Rajpurkar2018Know,lai2017race}. Although recent systems have reported human-parity performance on various benchmarks \cite{zhang2019sg,back2020neurquri,zhang2020retrospective} such as SQuAD \cite{rajpurkar-etal-2016-squad,Rajpurkar2018Know} and RACE \cite{lai2017race}, whether the machine has necessarily achieved human-level understanding remains controversial \cite{zhang2020machine,sugawara2021benchmarking}. 
Recently, there is increasing interest in improving machines' logical reasoning ability, which can be categorized into symbolic approaches and neural approaches. Notably, analytical reasoning machine (AMR) \cite{2021arXiv210406598Z} is a typical symbolic method that injects human prior knowledge to deduce legitimate solutions. 

\subsection{Logical Reasoning}
Neural and symbolic methods have been studied for logical reasoning \cite{garcez2015neural,besold2017neural,chen2019neural,ren2020beta,zhang2021video}. Compared with the neural methods for logical reasoning, symbolic approaches like \cite{wang2021logic} rely heavily on dataset-related predefined patterns which entails massive manual labor, greatly reducing the generalizability of models. Also, it could introduce propagated errors since the final prediction depends on the intermediately generated functions. Even if one finds the gold programs, executing the program is quite a consuming work as the search space is quite large and not easy to prune. Therefore, we focus on the neural research line in this work, to capture the logic clues from the natural language texts, without the rely on human expertise and extra annotation. 

Since the logical reasoning MRC task is a new task that there are only a few latest studies, we broaden the discussion to scope of the related tasks that require reasoning, such as commonsense reasoning \cite{davis2015commonsense,bhagavatula2019abductive,talmor2019commonsenseqa,huang2019cosmos},  multi-hop QA \cite{yang2018hotpotqa} and dialogue reasoning \cite{mutual}. Similar to our approach of discovering reasoning chains between element discourse and key phrases,  \citet{fang2019hierarchical} proposes a hierarchical graph network (HGN) that helps to multi-hop QA. Our method instead avoids the incorporation of external knowledge and designs the specific pattern for logical reasoning. Discourse-aware graph network (DAGN) proposed by \citet{zhang2021video} also uses discourse relations to help logical reasoning. However, only modeling the relation between sentences will ignore more fine-grained information. Focal Reasoner proposed by \citet{ouyang2021fact}, covering global and local knowledge as the basis for logic reasoning, is also an effective approach. In contrast, our work is more heuristic and has a lighter architecture.

Previous approaches commonly consider the entity-level, sentence-level relations, or heavily rely on external knowledge and fail to capture important interaction information, which are obviously not sufficient to solve the problem \cite{qiu2019dynamically,ding2019cognitive,chen2019multi}. Instead, we take advantages of inter-sentence EDUs and intra-sentence KPHs, to construct hierarchical interactions for reasoning. The fine-grained holistic features are used for measuring the logical fitness of the candidate answers and the given context. As our method enjoys the benefits of modeling reasoning chains from riddled texts, our model can be easily extended to other types of reasoning and inference tasks, especially where the given context has complex discourse structure and logical relations, like DialogQA, multi-hop QA and other more general NLI tasks. We left all the easy empirical verification of our method as future work.

\section{Conclusion}
This paper presents a novel method to guide the MRC model to better perform logical reasoning tasks. We propose a holistic graph-based system to model hierarchical logical reasoning chains. To our best knowledge, we are the first to deal with context at both discourse level and phrase level as the basis for logical reasoning. To decouple the interaction between the node features and type features, we apply hierarchical interaction mechanism to yield the appropriate representation for reading comprehension. On the logical QA benchmarks (ReClor, LogiQA) and natural language inference benchmarks (SNLI and ANLI), our proposed model has been shown effective by significantly outperforming the strong baselines.

\bibliography{anthology}
\bibliographystyle{acl_natbib}

\appendix
\clearpage

\section{KPHs Extraction Algorithm}
\label{kph_algorithms}
\begin{algorithm}[H]
\small
\caption{Key Phrases (KPH) Extraction Algorithm}\label{alg}
\begin{algorithmic}[1]
\Require Input $C=\{S^1, S^2, \dots, S^I \}$, $n$-gram length $n$, min word length $m$, number of Key Phrases $k$
\Ensure Set of Key phrases with top-$k$ TF-IDF weights $K =\{g_1, g_2, \dots, g_k\}$
\State Obtain the TF-IDF dictionary $\mathcal{F}$ = TF-IDF($C$)
\State Generate $n$-gram dictionary $G$ = \textsc{$n$-gram}($C, n$)
\State Filter $n$-gram dictionary $G$, $\tilde{G}$=$\textsc{Filter}(G,m)$
\State Retrieve the original expressions $K$ = \textsc{Retrieve}($C, \mathcal{F}, \tilde{G}, k$)
\Procedure{TF-IDF}{$C$}
\For{each sentence in $C$}
\State Filter stop-words in the sentence
\State Calculate the TF-IDF weight for each word
\EndFor
\State \textbf{return} TF-IDF dictionary $\mathcal{F}$
\EndProcedure
\Procedure{$n$-gram}{$C, n$}
\For{each sentence in $C$}
\State Select all gram $g$ with length $n$ in the sentence
\State Add $g$ to the dictionary $G$
\EndFor
\State \textbf{return} $n$-gram dictionary $G$
\EndProcedure
\Procedure{Filter}{$G, m$}
\For{each $n$-gram $g$ in $G$}
\If{stopwords in $g$ or length$(g)$ is less than $m$ or there is any number in $g$}
    \State Delete $g$
\EndIf
\If{length$(g)$ is 1 and POStag($g$) is not noun, verb, or adjective}
    \State Delete $g$
\EndIf
\EndFor
\State \textbf{return} $n$-gram dictionary $\tilde{G}$
\EndProcedure%
\Procedure{Retrieve}{$C, \mathcal{F}, \tilde{G}, k$}
\For{each $g$ in $\tilde{G}$}
\State Calculate the sum of the TF-IDF weights of each word in $g$, add to a dictionary $z = \{g: w(g)\}$
\EndFor
\State Rank the top-$k$ $n$-gram $g$ by TF-IDF weight sum. Construct key phrases set $K =\{g_1, g_2, \dots, g_k\}$
\If{$n$=1}
    \For{each $g$ in $K$}
    \State $g_s$= STEM($g$)
    \State Retrieve all the original words from $C$ containing $g_s$, add to $K$
\EndFor\EndIf
\State \textbf{return}  Set of Key phrases $K =\{K_1, K_2, \dots, K_k\}$
\EndProcedure%
\end{algorithmic}
\end{algorithm}

\section{Dataset Information}\label{dataset}
\paragraph{ReClor} The Reading Comprehension dataset requiring logical reasoning (ReClor) is extracted from standardized graduate admission examinations \cite{yu2020reclor}. It contains 6,138 multiple-choice questions modified from standardized tests such as GMAT and LSAT and is randomly split into train/dev/test sets with 4,638/500/1,000 samples respectively. Multiple types of logical reasoning question are included.
\paragraph{LogiQA} LogiQA is sourced from expert-written questions for testing human Logical reasoning. It contains 8,678 QA pairs, covering multiple types of deductive reasoning. It is randomly split into train/dev/test sets with 7,376/651/651 samples respectively.
\paragraph{SNLI}  The Stanford Natural Language Inference (SNLI) dataset contains 570k human annotated sentence pairs, in which the premises are drawn from the captions of the Flickr30 corpus and hypotheses are manually annotated. The full dataset is randomly split into 549k/9.8k/9.8k. This is the most widely used entailment dataset for natural language inference. It requires models to take a pair of sentence as input and classify their relation types, i.e., \textsc{entailment},\textsc{neutral}, or \textsc{contradiction}.
\paragraph{ANLI} The Adversarial Natural Language Inference (ANLI) is a new large-scale NLI benchmark dataset, collected via an iterative, adversarial human-and-model-in-the-loop procedure. Specifically, the instances are chosen to be difficult for the state-of-the-art models such as BERT and RoBERTa. A1, A2 and A3 are the datasets collected in three rounds. A1 and A2 are sampled from Wiki and A3 is from News. It requires models to take a set of context, hyperthesis and reason classify the label (\textsc{entailment},\textsc{neutral}, or \textsc{contradiction}). A1 has 18,946 in total and is split into 16,946/1,000/1,000. A2 has 47,460 in total and is split into 45,460/1,000/1,000. A3 has 102,859 in total and is split into 100,459/1,200/1,200. ANLI refers to the combination of A1, A2 and A3.
\begin{figure*}[t]
    \centering
    \includegraphics[width=1.0\textwidth]{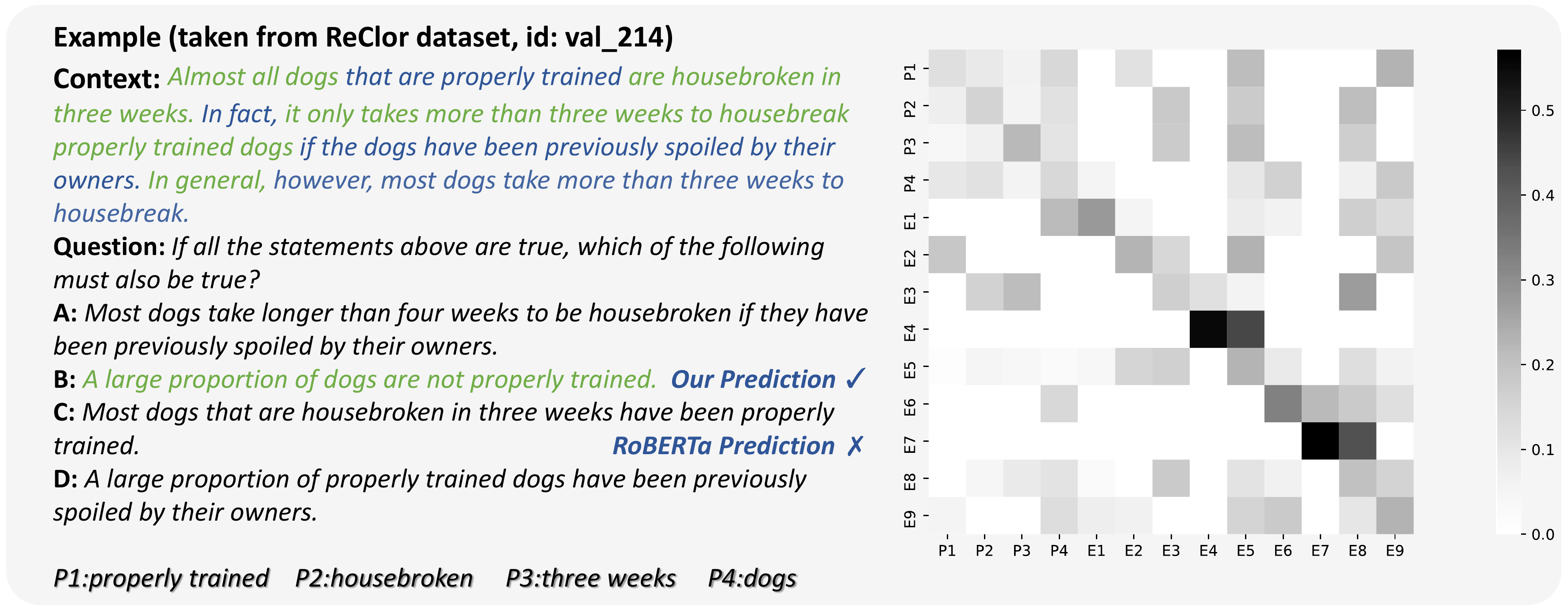}
    \caption{An example showing the logical reasoning capability of our model (Left) and the corresponding attention map (Right). EDUs are shown in different colors alternately, corresponding to E1-E9 in the attention map.}
    \label{case}
\end{figure*}
\section{Parameter Selection}\label{params}
Our model is implemented based on the Transformers Library \cite{wolf-etal-2020-transformers}. Adam \cite{2014arxiv1412.6980k} is used as our optimizer. The best threshold for defining semantic relevance is 0.5.
We run 10 epochs for ReClor and LogiQA, 5 epochs for SNLI and ANLI, and select the model that achieves the best result in validation. Our models are trained on one 32G NVIDIA Tesla V100 GPU. The training time is around half an hour for each epoch. The maximum sequence length is 256 for ReClor and SNLI, 384 for LogiQA and 128 for ANLI. The weight decay is
0.01. We set the warm-up proportion during training to 0.1. We provide training configurations used across our experiments in Table~\ref{parameters}.

\begin{table}
\centering
\setlength{\tabcolsep}{2.3pt}{
\begin{tabular}{llcc}
\toprule
Dataset&PrLM & batchsize&learning rate\\
\hline
\multirow{3}{*}{ReClor}&RoBERTa-base& 24 &1e-5\\
&RoBERTa-large& 32 &8e-6 \\
&DeBERTa-xlarge& 8 &8e-6\\
\midrule
\multirow{3}{*}{LogiQA}&RoBERTa-base& 2 &4e-6\\
&RoBERTa-large& 2 &8e-6 \\
&DeBERTa-xlarge& 2 &8e-6\\
\midrule
\multirow{2}{*}{SNLI}&BERT-base& 32 & 2e-5\\
&RoBERTa-large& 32 &2e-5 \\
\midrule
ANLI&RoBERTa-large& 32 &2e-05 \\
\bottomrule
\end{tabular}}
\caption{Parameter Selection\label{parameters}}
\end{table}

\section{Case Study}\label{appendix:case}
To intuitively show how our model works, we select an example from ReClor as shown in Figure \ref{case}, whose answer is predicted correctly by our model but not by baseline models (RoBERTa). The example shows that powerful pre-trained language models such as RoBERTa may be better at dealing with sentence pairs that contain overlap parts or similar words. For example, the wrong answer chosen by RoBERTa is another expression of the first sentence in the given context. The words are basically the same, only the order changes. The model itself does not understand the logical relation between sentences and phrases, but only compares their common elements for prediction, failing in logical reasoning task. In contrast, our model can not only match synonymic expressions, but also make logical inferences  by separating sentences into EDUs and extracting key phrases and establishing logical relations between them. The importance of those elements are interpreted by the attention distribution as shown in the right part, which is derived from the last layer of our model. 

\end{document}